\documentclass[fleqn,10pt,final]{wlscirep} 
\usepackage[utf8]{inputenc}
\usepackage[T1]{fontenc}
\usepackage{graphicx}

\usepackage{hyperref}
\usepackage[printonlyused]{acronym}
\usepackage{changes}
\definechangesauthor[name={Greg}, color=blue]{GP}
\definechangesauthor[name={Jeff}, color=purple]{JR}
\definechangesauthor[name={Soeren}, color=red]{SR}
\definechangesauthor[name={Yandong}, color=red]{YD}
\usepackage{dirtytalk}
\usepackage[justification=centering]{caption}
\definecolor{black}{rgb}		{0.0, 0.0, 0.0}
\definecolor{white}{rgb}		{1.0, 1.0, 1.0}
\definecolor{yellow}{rgb}		{1.0, 1.0, 0.8}
\definecolor{red}{rgb}			{0.6, 0.0, 0.2}
\definecolor{blue}{rgb}		{0.0, 0.2, 0.5}
\definecolor{green}{rgb}		{0.6, 0.8, 0.8}
\definecolor{dark_green}{RGB} {0, 140, 0}
\definecolor{gold}{rgb}		{0.6, 0.4, 0.1}
\definecolor{grey}{RGB}{0,0,0}
\definecolor{Gray}{gray}{0.8}
\definecolor{MediumGray}{gray}{0.9}
\definecolor{LightGray}{gray}{0.98}
\definecolor{LightCyan}{rgb}{0.88,1,1}
\definecolor{purple}{RGB}{128,0,128}
\definecolor{sl_blue}{RGB}{47, 60, 105}
\definecolor{orange}{RGB}{255,165,0}
\definecolor{Gray}{gray}{0.85}

\newcommand{\eg}{e.\,g.,\ }
\newcommand{\ie}{i.\,e.,\ }
\newcommand{\wrt}{w.\,r.\,t.\ }

\usepackage{xspace}
\newcommand{\knockonerror}{knock-on error\xspace}
\newcommand{\knockonerrors}{knock-on errors\xspace}

\title{Identifying Cause-and-Effect Relationships of Manufacturing Errors using Sequence-to-Sequence Learning}

\author[1,+,*]{Jeff Reimer}
\author[1,+]{Yandong Wang}
\author[1,]{Sofiane Laridi}
\author[3,]{\added[id=JR]{Juergen Urdich}}
\author[2,]{Sören Wilmsmeier}
\author[1,]{Gregory Palmer}

\affil[1]{L3S Research Center, Leibniz University of Hanover, 30167 Hanover, Germany}
\affil[2]{Institut für Fertigungstechnik und Werkzeugmaschinen, Leibniz University of Hanover, 30167 Hanover, Germany}
\affil[3]{\added[id=JR]{Volkswagen Group, Post Office Box 21 05 80, 30405 Hanover, Germany}}

\affil[*]{reimer@l3s.de}

\affil[+]{these authors contributed equally to this work}

\begin{abstract}
In car-body production the pre-formed sheet metal parts of the body 
are assembled on fully-automated production lines. 
The body passes through multiple stations in succession, 
and is processed according to the order requirements. 
The timely completion of orders depends on the individual station-based operations concluding within their scheduled cycle times. 
If an error occurs in one station, it can have a knock-on effect, resulting in delays on the downstream stations. 
To the best of our knowledge, there exist no methods for automatically distinguishing between \emph{source} and \emph{knock-on} errors in this setting, as well as establishing a causal relation between them.
Utilizing real-time information about conditions collected by a production data acquisition system, 
we propose a novel \emph{vehicle manufacturing analysis system}, which uses deep learning to establish a link between source and knock-on errors.
We benchmark three sequence-to-sequence models, and introduce a novel composite time-weighted action metric for evaluating models in this context.
We evaluate our framework on a real-world car production dataset recorded by \replaced[id=JR]{Volkswagen Commercial Vehicles}{a large German car manufacturer}.
\added[id=GP]{Surprisingly we find that \replaced[id=GP]{71.68}{75}\% of sequences contain either a source or knock-on error.} 
\added[id=GP]{With respect to seq2seq model training, we find that the Transformer demonstrates a better performance compared to LSTM and GRU in this domain, in particular when the prediction range with respect to the durations of future actions is increased.}

\end{abstract}
\begin{document}

\flushbottom
\maketitle
%
%
\thispagestyle{empty}


\section*{Introduction} \label{sec:intro}

Time-series forecasting is increasingly being used for predicting future events
within business and industry to enable informed decision making~\cite{CHAN16,ZHAO14}.
In this paper we evaluate its potential to revolutionize automated vehicle manufacturing, where real-time information is collected by a \emph{production data acquisition} (PDA) system. 
The effects of errors in interlinked manufacturing systems have dire consequences, such as production delays and even production system failure. 
In industrial manufacturing, downtimes are associated with high costs. 
To counteract downtimes, research and development has so far focused on 
predictive maintenance of equipment~\cite{arena2022predictive} and the use of alternative
manufacturing routes through the production process~\cite{denkena2021scalable}. 
However, these approaches do not explicitly focus on delays (micro-disturbances) in individual process steps,
which are propagated throughout the process chain and amplified in the process.

%
%
%
The optimal utilisation of a fully automated car body production line depends on the individual station-based work steps completing within their scheduled cycle times. 
%
%
However, various disturbances with statistical significance are often detected. 
In particular, \emph{source errors} (typically logged by the PDA system, e.g., \say{No components available.}), may not only impact the current station,
but also have a detrimental effect on the downstream workstations (hereinafter referred to as \emph{stations}), resulting in \emph{\knockonerrors} and delays. 
Even minimal delays that are barely noticeable by humans can result in high additional costs.

While time-series forecasting for car-body production is challenging (due to discontinuities, spikes and segments \cite{FRAU15}), 
deviations in the manufacturing process can be identified through comprehensive production data acquisition and the structured evaluation of these data . 
However, currently process delays and anomalies are identified through 
rule based classifiers that are manually programmed and maintained using extensive domain knowledge. 
In addition, further efforts are incurred in the interpretation of the processed data. 
This prevents production staff from rapidly deploying targeted countermeasures.

To the best of our knowledge no approach currently exists that automatically: i.)~learns to classify both source and \knockonerrors; ii.)~establish a link
between errors; and iii.)~measures the knock-on effect of source errors. In this work we take steps towards solving these challenges using machine learning (ML). 
%

Our contributions can be summarized as follows: 
%

\noindent{\textbf{i.)}} We introduce an ML-based \emph{vehicle manufacturing analysis system} (VMAS) for process monitoring and cycle time optimization.
The system is designed to detect delays and malfunctions in the production process early and automatically without manual effort. 
Furthermore, it identifies cause-effect relationships and predicts critical errors using sequence-to-sequence (seq2seq) models.

\noindent{\textbf{ii.)}} To enable a fair comparison between different seq2seq architectures for predicting errors in this context, we introduce a novel \emph{Composite Time-weighted Action} (CTA) metric.
Our metric allows stakeholders to weight the sequences of predictions output by our model, and choose to what extent immediate action duration predictions are prioritized over distant ones.

\noindent{\textbf{iii.)}} Our VMAS is evaluated on PDA system data from the car body production of \replaced[id=JR]{Volkswagen Commercial Vehicles}{a large German car manufacturer}. This includes the benchmarking of a number of popular seq2seq models for learning cause-effect relationships, including LSTM, GRU and Transformer. Surprisingly our evaluation shows the prevalence of source and \knockonerrors, which occur in \replaced[id=YD]{71.68\%}{ 75\%} of action sequences.
The evaluation of prediction component meanwhile shows that 
the Transformer outperforms LSTM and GRU models, 
capable of accurately predicting the durations of up to seven actions into the future.

\section*{Problem Definition} \label{sec:problem_definition}

\begin{figure}[h]
    \centering
    \includegraphics[width=0.8\textwidth]{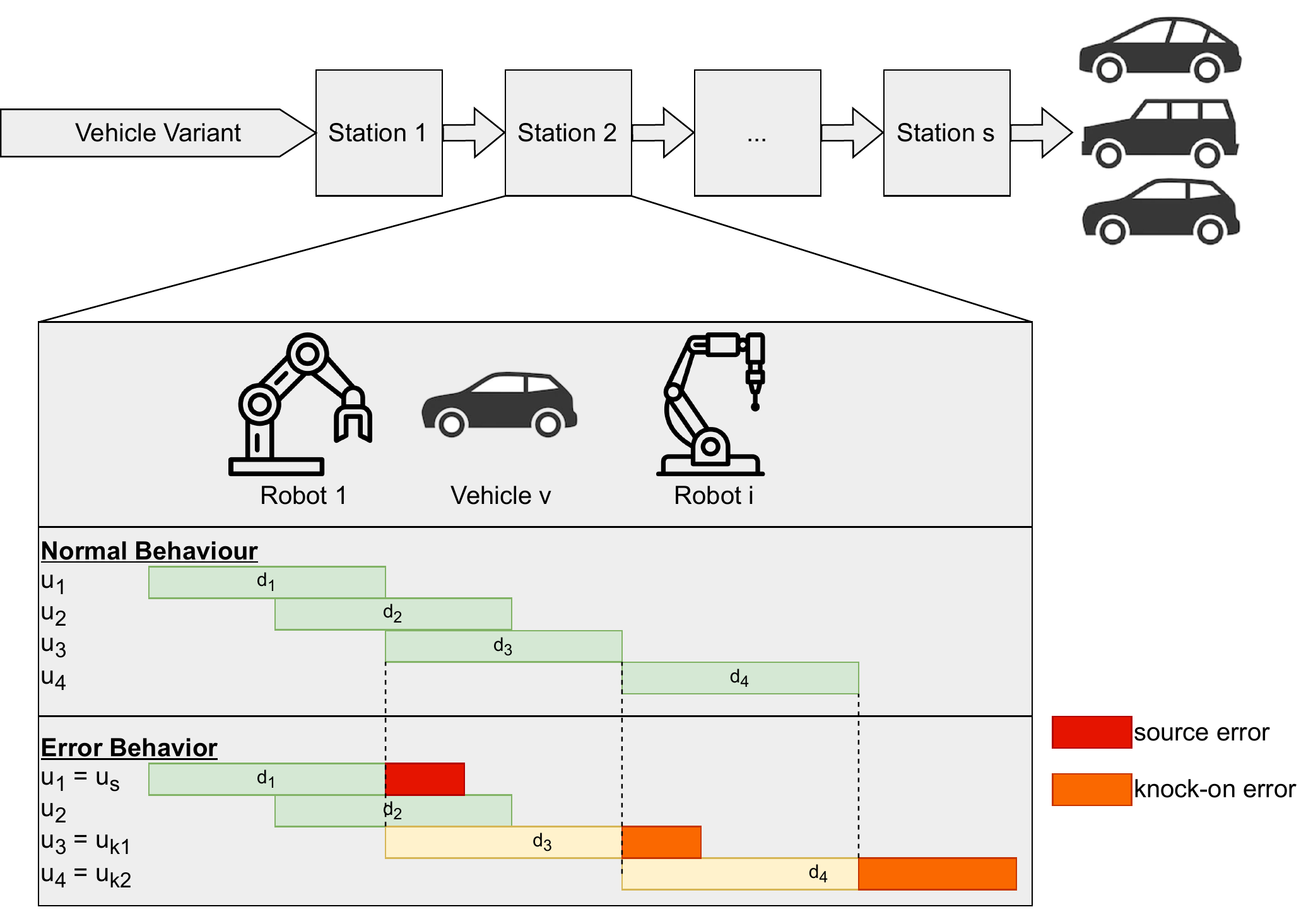}
    \caption{Production line for assembling vehicle variants, illustrating normal behaviour and error behavior in a Gantt-chart, with respect to durations $d$. Several actions can happen in parallel. In the error behavior scenario a source error for an action duration tuple $u_s$ (marked red) can lead to multiple \knockonerror $u_{k1}$ and $u_{k2}$ (marked orange).}
    \label{AssemblyLine}
\end{figure}

In car manufacturing, the vehicle body is processed through visiting a sequence of fully-automated stations.
Each station comprises an ensemble of manufacturing robots (see Figure~\ref{AssemblyLine}).
The stations are clocked out to measure the timeliness 
of the 
vehicles to-be assembled until they exit the production line. 
At each station actions are performed, which we define as a triple
$a = (s, v, i)$,
consisting of a station $s$, 
vehicle code $v$, and action~ID~$i$. 
The production of vehicles also includes variants (left or right-hand drive vehicles for example) and therefore the nominal action is variable dependent on the vehicle variant, which is information included in the vehicle code.

Each action describes a specific and accomplished production step (for example transportation or manufacturing step).
We are interested in the duration $d$ required to complete each executed action $a$,
which can be viewed as an action duration tuple $u=(s, v, i, d)$. 
For notational convenience we shall refer to $d^a$ as the duration taken by an action $a$.
%
%
%
In a clocked out vehicle production system, for each action
$a$ there exists an expected maximum allowed duration $d^a_{max}$.
The duration of an action $a$ must therefore be less than, or equal to, this expected allowed maximum time: $d^a \leq d^a_{max}$.
In this work, we focus on sequences of actions and their durations, 
i.e., chains of action\replaced[id=GP]{ duration tuples}{s}, defined as $x = (u_1, u_2, ... , u_{n})$. 
%
It is worth noting however, that actions can overlap, e.g., be executed in parallel.
Therefore, it is not the case that one particular action
has to have completed its task before another action can start.
The sequence of actions is also dependent on the vehicle variant.

Malfunctions are a recurring problem in production.
In the rare instance that a malfunction causes a long period of downtime, usually a situation analysis is conducted and possible fix is performed by staff engineers in the factory.
However, our focus is on the small, seemingly insignificant and common delays, 
that not only have an effect on a station itself, but where subsequent perturbation propagate to downstream stations, causing further delays.
Here we consider executed actions with two types of errors resulting from delays, where the duration 
$d^a > d^a_{max}$: 
i.) \emph{source errors}, $u_s$ where an abnormal action duration is accompanied by an error message;
ii.) \emph{\knockonerrors}, where an action $u_k$ with an abnormally long action duration is not accompanied by an error message. 
In this work we are interested in knock-on errors that occur after a source error (\ie a logged error) within the sequence of actions: $(\text{..., } u_{s} \text{, ..., } u_{k} \text{, ...})$. 

An individual source error may appear inconspicuous, since source errors do not have to deviate significantly from the normal time. 
However, the \knockonerrors, which also do not have to deviate much individually, can result in a significant accumulated time-delay. 
From the PDA system it is not possible to understand the scope of downstream actions and the knock-on effects of a source error.
%
It is only possible to assert that downstream actions can accumulate time-delays without reported fault messages.
Consequentially, this leads to a significant loss of effective production time overall.

The analysis of the relationship between source and \knockonerrors is challenging due to the latent entanglement of the individual processes of actions. 
\deleted[id=JR]{Therefore, in addition to source and \knockonerrors we define 
\emph{pattern errors}, as an arrangement between source and subsequent propagating \knockonerrors in the sequence of actions.} 
\deleted[id=JR]{Errors must show a high enough frequency that is statistically significant to be considered pattern errors.}
An argument can be made that a rule-based model can determine the relationship of a \replaced[id=JR]{source and knock-on errors}{pattern error}.
However, this approach requires extensive domain knowledge and the resulting model would not be transferable across stations.
We hypothesize that deep learning-based seq2seq models are able to learn the nominal sequence of actions and, more importantly for the producer, the \replaced[id=JR]{recurring source and knock-on}{pattern} errors in them as well.
If the \deleted[id=JR]{pattern}errors can be predicted with a satisfying accuracy, then it means inherent causal-effect rules are learned from the abundance of data.

\section*{Related Work} \label{sec:relevant_work}


Within the context of intelligent industrial production a significant amount of research has been dedicated towards forecasting, failure prediction and anomaly detection using time series data~\cite{DING20,JIAN20}.
The literature in this area provides an overview of the suitability of approaches designed to solve these problems when applied to various production contexts, often featuring a comparison between traditional machine learning approaches and that of advanced deep neural networks.
Failure prediction for instance has often been limited to standard key performance indicators.
Moura et al.~\cite{Mour11} evaluate the effectiveness of support vector machines in forecasting time-to-failure and reliability of engineered components based on time series data.
Yadav et al.~\cite{Yada12} present a procedure to forecast time-between-failure of software during its testing phase by employing fuzzy time series approach.
Others use artificial neural networks or statistical approaches to model machine tool failure durations continuously and cause-specific \cite{Denk20, Olad06}.

\emph{Recurrent neural network} (RNN) models meanwhile are capable of identifying long-term dependencies from time-series data directly~\cite{chung2014empirical}.
Successes here include: multi-step time-series forecasting of future system load with the goal of performing anomaly detection and system resource management, enabling the automated scaling in anticipation of changes to the load~\cite{Jero18}; and using stacked LSTM networks to detect deviations from normal behaviour without any pre-specified context window or pre-processing~\cite{Malh15}.
However, the 
performance of 
encoder-decoder architectures relying on memory cells alone typically suffers, as the encoding step must learn a representation for an (potentially lengthy) input sequence. 
Here \emph{attention based encoder-decoder architectures} provide a solution, where the hidden states from all encoder nodes are made available at every time step.
In-fact, pioneering work by \cite{VASW17} demonstrated that one can dispense with recurrent units and rely solely on the attention, introducing the Transformer.
%
Further improvements can be obtained via Transformers implemented with GRUs~\cite{parisotto2020stabilizing}.

Not surprisingly attention based approaches are increasingly being applied to industry problems~\cite{9536749}.
Li et al.~ \cite{LI21} present a novel approach to extracting dynamic time-delays to reconstruct multivariate data for an improved attention-based LSTM prediction model and apply it in the context of industrial distillation and methanol production processes.
But they do not explicitly consider failure propagation in concatenated manufacturing systems to evaluate failure criticality and to generate a reliable failure impact prediction. 
%
Attention-based models have also been applied to failure prediction and rated as favorable.
LI et al.~\cite{LI22} propose an attention-based deep survival model to convert a sequence of signals to a sequence of survival probabilities in the context of real-time monitoring.
While Jiang et al.~\cite{JIAN20} use time series multiple channel convolutional neural network integrated with the attention-based LSTM network for remaining useful life prediction of bearings.
Near real-time disturbance detection becomes possible with the attention-based LSTM encoder– decoder network by Yuan et al.~\cite{YUAN20}, which allows to align an input time series with the output time series and to dynamically choose the most relevant contextual information while forecasting.
%
In contrast to previous work, we propose a workflow and  evaluate seq2seq approaches for failure impact prediction in concatenated manufacturing systems.

\section*{Vehicle Manufacturing Analysis System}



\begin{figure}[h]
\centering
\includegraphics[width=0.8\textwidth]{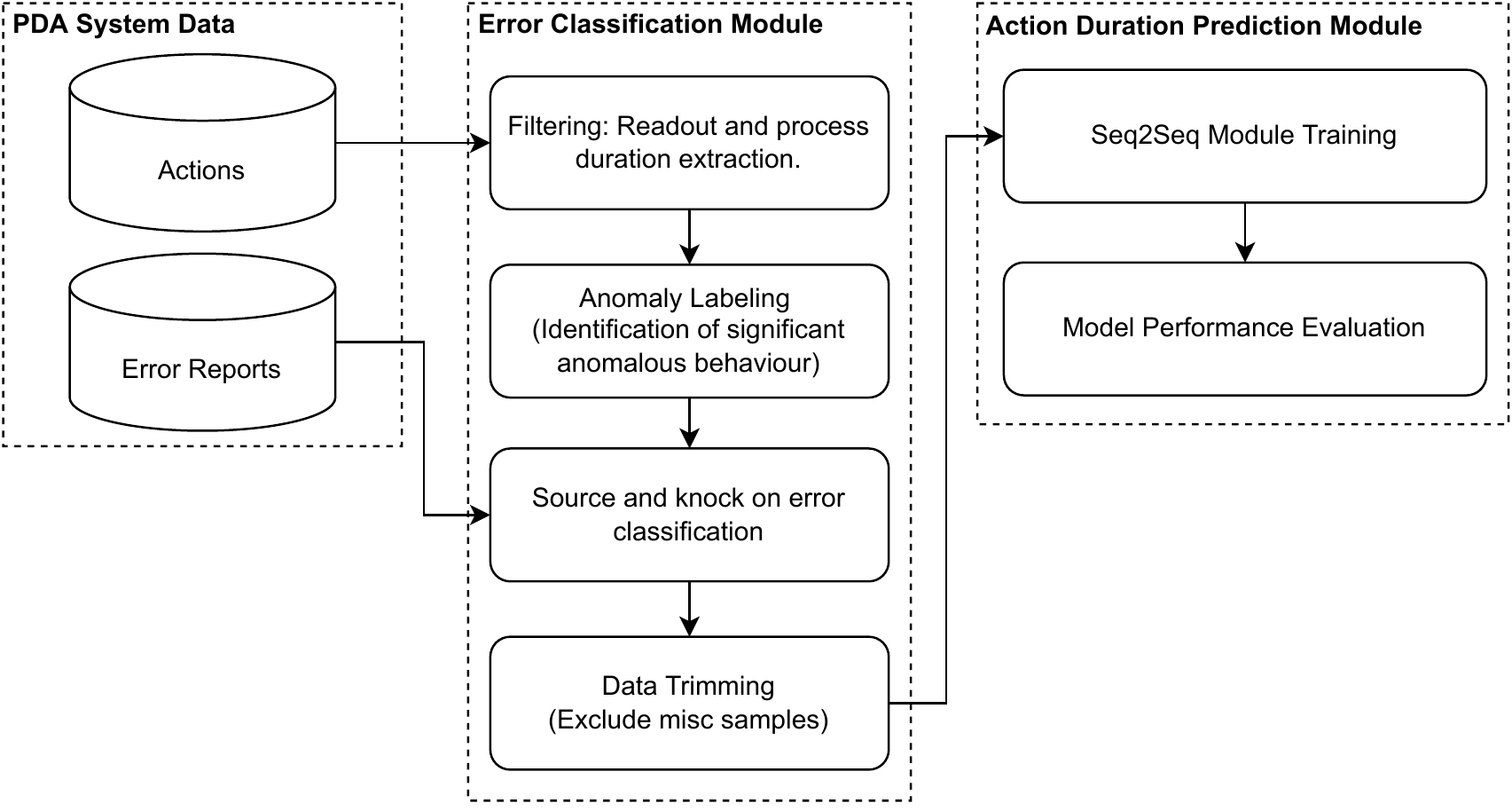}
\caption{Flowchart of our vehicle manufacturing analyses system (VMAS).
First the PDA system data is processed by our Error Classification Module, 
resulting in four sub-sets: source errors, \knockonerrors, normal and misc. 
The resulting source and \knockonerror sets can then be used by 
our stakeholders for obtaining valuable insights \wrt causes of 
delays.
Next, upon excluding misc samples, we use our data for training 
sequence-to-sequence models
for predicting future delays. 
%
}
\label{Flowchart}
\end{figure}

In this section we introduce our vehicle manufacturing analyses system (VMAS), which we developed according to the cross-industry standard process for data mining (CRISP-DM) \cite{chapman1999crisp}. 
%
%
Our use-case has two separate databases that store \emph{cycle times} and \emph{error reports} data respectively. 
%
The PDA system in our use-case registers 
%
and stores action duration tuples~$u$ in the cycle times database.
%
%
The data are processed by our VMAS, which consists of two main components: 1.) an error classification module for identifying source and \knockonerrors within our dataset; and 2.) a duration prediction module, trained to predict the time required for $n$ future actions. We describe each component in detail below and a flowchart can be found in Figure~\ref{Flowchart}. 

\subsection*{Module 1: Error Classification} 

We begin with an actions dataset $\mathcal{D}_a$ and an error reports database that stores timestamped error logs as well as the duration of the logged errors.
Each sample $x \in \mathcal{D}_a$, is a sequence of action duration tuples $x = (u_0, u_{1}, u_{2}, \text{...,} u_n)$, where $n$ is the number of actions executed during a \emph{complete sequence}.
The error classification module of our workflow allows us to identify
the most significant errors within our dataset,
and distinguishes source from \knockonerrors.
More specifically, this module allows us to split samples from
our dataset into four subsets: {normal} $\mathcal{D}_{n}$,
{source errors} $\mathcal{D}_{s}$, {\knockonerrors} $\mathcal{D}_{k}$ and {misc} $\mathcal{D}_{m}$.
This splitting of the dataset into sub-sets serves two 
purposes: 
i.)~The classification in $\mathcal{D}_{s}$ and $\mathcal{D}_{k}$ helps the stakeholder to conduct an automated analysis of all actions and it eliminates the need for manual and often time consuming inspection of actions;
ii.)~During preliminary trials we found that samples from $\mathcal{D}_{m}$ are exceedingly rare and disturb the training of the seq2seq models. 
Therefore, the error classification module also provides
a valuable preprocessing step prior to training our 
seq2seq models to predict future delays.
Below we first discuss our approach for labelling
our samples,
and then formally define the conditions for a sequence
$x$ to belong to one of the four subsets. 
We note that for our VMAS there is an assumption that \emph{all} source errors are \emph{logged errors}.

\begin{figure}[h]
    \centering
    \includegraphics[width=0.8\textwidth]{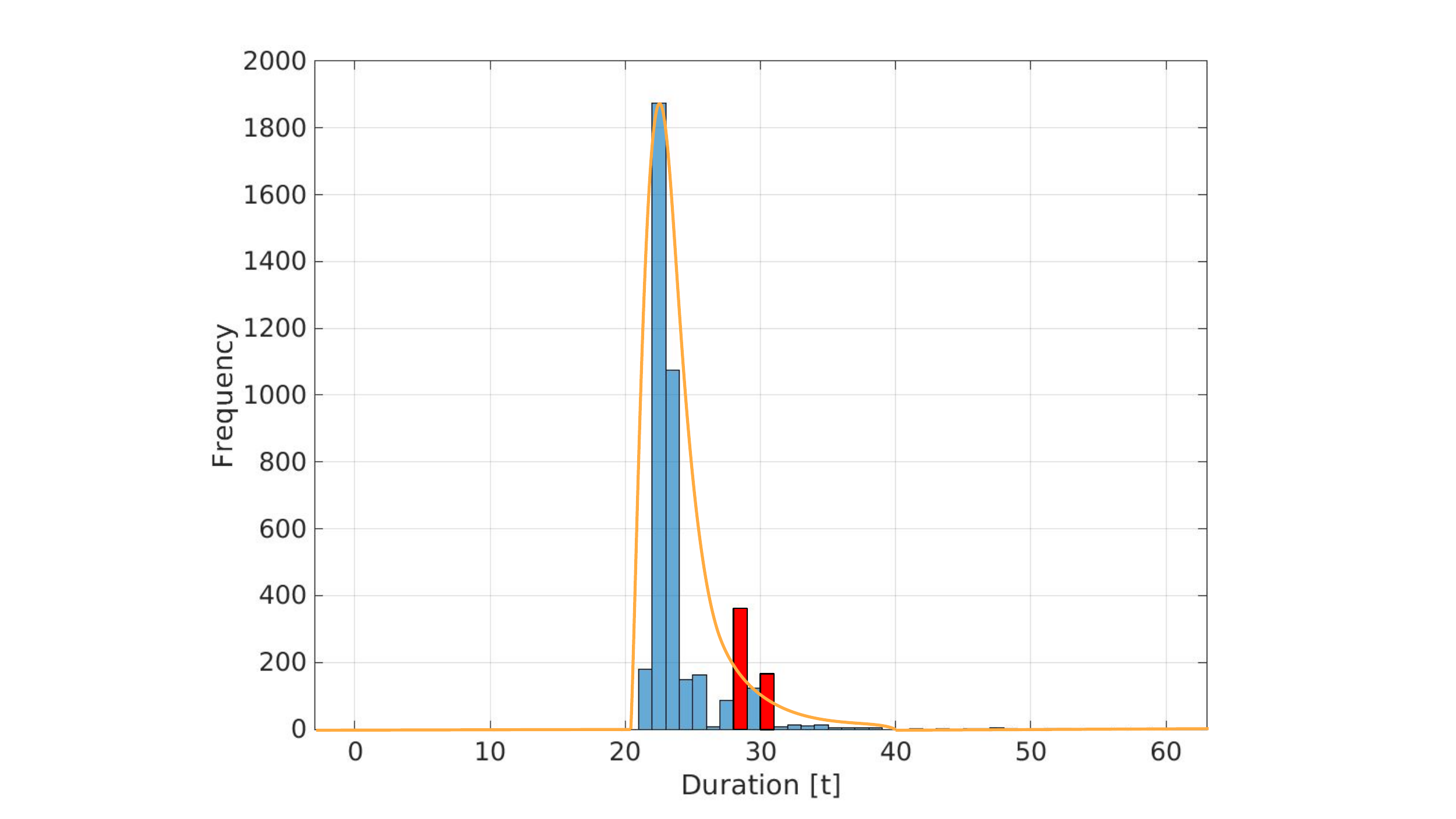}
    \caption{Peak detection in a histogram of durations for a specific action utilizing MLE as a threshold; peaks above the MLE threshold (red) are considered as \replaced[id=JR]{significant}{pattern} errors.
    }
    \label{fig:peak_detection}
\end{figure}

\textbf{Labelling:}
We use the maximum likelihood estimation (MLE) method for the labelling of anomalous behavior.
%
For each action $a$, a 
normal (Gaussian) distribution is sought that fits the existing data distribution
with respect to the frequency of each duration (for an example see Figure~\ref{fig:peak_detection}).
%
%
The density function of the normal distribution contains two parameters: the expected value  $\mu$  and standard deviation $\sigma$, which determine the shape of the density function and the probability corresponding to a point in the distribution. 
The MLE method is a parametric estimation procedure that finds $\mu$ and $\sigma$ that seem most plausible for the distribution of the observation $z$:
\begin{equation} \label{eq:4.1_from_original}
f(z \mid \mu, \sigma^2) = \frac{1}{\sqrt{2\pi\sigma^2}} \exp\left(-\frac{(z-\mu)^2}{2\sigma^2}\right).    
\end{equation}
The density function describes the magnitude of the probability of $z$ coming from a distribution with $\mu$ and $\sigma$.
The joint density function can be factorised as follows: 
\begin{equation} \label{eq:4.2_from_original}
f(z_1, z_2, ..., z_n \mid \vartheta) = \Pi^n_{i=1}f(z_i \mid \vartheta)
\end{equation}
For a fixed observed variable, the joint density function of $z$ can be interpreted.
This leads to likelihood function: 
\begin{equation} \label{eq:4.3_from_original}
L(\vartheta) = \Pi^n_{i=1}f_\vartheta(z_i)
\end{equation}
The value of $\vartheta$ is sought for which the sample values $z_1, z_2, ..., z_n$ have the largest density function.
Therefore, the higher the likelihood, the more plausible a parameter value $\vartheta$ is.
As long as the likelihood function is differentiable, the maximum of the function can be determined.
Thus, the parameters $\mu$ and $\sigma$ can be obtained.

Next, we seek to identify \emph{high frequency peaks} with
respect to the durations $d^a$ for an action $a$, that exceed the nominal duration $d^a_{norm}$. 
We are interested in significant errors, where we use the MLE
threshold to determine if an error is significant or not.
We denote significant errors as $d^a_{sig}$.
These abnormal and distinct duration are indicating \replaced[id=JR]{a recurring behaviour}{the recurring pattern error behaviour}.
%
%
We formally define the criteria for each sub-set below:

\begin{itemize}
    \item  \textbf{Source errors}
are
samples where for each complete sequence $x$, 
we have at least one action duration that is considered critical, 
of statistical significance, and is accompanied by
an error message. More formally:
%
a complete action sequence $x$ is considered a \emph{source
error sequence} $x \in \mathcal{D}_{s}$ iff there exists an action duration tuple $u \in x$, where the duration is $d^a_{sig}$ and there is a corresponding error message in the error reports database.

\item \textbf{Knock on errors} 
meet the same criteria as source errors, 
but lack an accompanying error message for $d^a_{sig}$.
%
Therefore, a complete action sequence $x$ is considered a \emph{\knockonerror sequence} $x \in \mathcal{D}_{s}$ iff there exists an action duration tuple $u \in x$, where the duration is $d^a_{sig}$ and there is not a corresponding error message in in the error reports database. 

\item \textbf{Normal} samples don't include $d^a_{sig}$.
Therefore, a complete sequence $x$ is considered a \emph{normal sequence} $x \in \mathcal{D}_{n}$ iff for all $u \in x$ there does not exist a duration $d^a_{sig}$. 

\item \textbf{Misc.} contains two types of complete action sequences:
i.) where for an action $u$ there is a duration $d^a_{sig}$ that is above a defined global threshold $d^a_{globalmax}$, meaning the duration is either intended (\eg the production line is paused), or staff are handling them; and ii.) where $x$ consists only of duration $d$ that exceed the nominal duration, but each of low significance, \ie not exceeding the corresponding MLE threshold.
%
\end{itemize}

It is worth noting that $\mathcal{D}_{n} \cup \mathcal{D}_{s} \cup \mathcal{D}_{k}$ 
may contain individual $d^a$ above the nominal duration, but below the threshold determined by the MLE, and therefore are errors of low significance. 
There can also exist an intersection between source and \knockonerrors.
Furthermore, the labelling of \knockonerrors is deliberately modular, as different 
methods can be applied here based on the stakeholder's requirements.
Naturally this will impact the subsequent
training of our seq2seq models, and 
therefore their predictions.

\subsection*{Module 2: Action Duration Prediction} \label{subsec:seq2seq}

While our error classification module assigns labels
to past errors, our second module focuses on the prediction
of future errors. 
Upon removing misc samples, we utilize our dataset to train seq2seq models to predict \knockonerrors. 
Given a sequence of action duration tuples our objective is to predict the time required by each of the next $n$ steps.
We therfore convert the data received from the error classification module into a dataset containing pairs $(x, y) \in \mathcal{D}$, 
where each $x$ is a sequence of action duration tuples $x = (u_{t-n}, u_{t-n+1}, u_{t-n+2}, \text{...,} u_t)$, and $y$ is the duration of the $n$ actions that follow 
$y = (d^a_{t}, d^a_{t+1}, d^a_{t+2}, ..., d^a_{t+n})$.
Using these data, we train and evaluate popular seq2seq models, including LSTM~\cite{hochreiter1997long}, GRU~\cite{chung2014empirical} and the Transformer~\cite{VASW17}. 
The later is of particular interest, as it represents the current state-of-the-art for a number of seq2seq tasks. 
Vaswani et al.~\cite{VASW17} presented the Transformer architecture for the Natural Language Processing (NLP) or Transductor task domain.
Previous RNN/CNN architectures pose a natural obstacle to the parallelization of sequences.
The Transformer architecture replaces the recurrent architecture by its attention mechanism and encodes the symbolic position in the sequence.
This relates two distant sequences of input and output, which in turn can take place in parallel.
The time for training is thereby significantly shortened.
At the same time, the sequential computation is reduced and the complexity $O(1)$ of dependencies between two symbols, regardless of their distance from each other in the sequence, remains the same~\cite{VASW17}.
Next we consider a novel metric for fairly evaluating models of different architectures -- in particular regarding the number of steps $n$ -- using a single scalar.
%



\section*{Composite Time-weighted Actions Metric}
%
\replaced[id=JR]{A sequence of actions can consist either of nominal behaviour or error behaviour by having at least one source or knock-on error included.}{A sequence of actions can either consist of nominal behaviour, or include a pattern error form distinct behaviours.}
To predict a distinct behavior we pass a partial sequence of actions to a seq2seq model to predict $n$ actions into the future.
However, in production there are a number 
of scenarios (including our current one), where a 
greater weighting needs to be placed on the performance
of the classifier with respect to short term predictions 
in order to enable a quick intervention.
%
Therefore, 
to evaluate our model in this setting a metric is required that: 
i.) assigns a higher importance on the immediate predictions versus later predictions in the sequence of actions; ii.) allows a prediction of quality invariant of the number of predicted future steps $n$, in order to cross compare various setups; iii.) has high precision when predicting the duration of an action.
For the evaluation of any seq2seq model we introduce the \emph{Composite Time-weighted Action} (CTA) metric.
The CTA is a convex combination of a Time-weighted Action RMSE (which we introduce below) and an F1 score that uses a threshold $b$:
%
%
\begin{equation}
\text{CTA} = \tau (\text{TARMSE})+ (1-\tau)(\text{F1}).     
\end{equation}
In the above equation stakeholders can use the weighting $\tau$ to either emphasize the TARMSE or precision when evaluating and comparing models. 
In the following we will discuss the two components.

\textbf{Time-weighted Action RMSE (TARMSE):}\label{TWA-RMSE}
To measure the performance of a model globally, we introduce a \emph{Time-weighted RMSE} that returns a single scalar metric for the $n$ model outputs. 
The model performance should not diminish if the starting point of predictions varies within the sequence of actions.
For our current problem setting immediate predictions should also have a higher importance than later ones.
%
In order to compensate for the increase of uncertainty we introduce a weighting factor $\beta_{i} = e^{-i}$ with $i$ being the action index.
%
%
The following formula is considering only predictions which are below the expected allowed maximum time $d^a_{globalmax}$:
%
%
\begin{equation}\label{eq:TARMSE}
\text{TARMSE}(n,k) =  \frac{1}{k\times S(n)} \sum_{i=1}^{n} \beta_{i} (k-R_{i})    
\end{equation}
 with
 \begin{equation}
 S(n) = \frac{e^{-n}-1}{1-e} = \sum_{i=1}^{n} \beta_{i} 
 \end{equation}
and
\begin{equation}
\beta_{i} = e^{-i}.
\end{equation}
In Equation~\ref{eq:TARMSE} $R_i$ is the RMSE for action $i$ and the $k$ value is oriented to the mean standard deviation of all the times of actions in this station within the max tolerance.
The standard deviation has the property of fitting a Gaussian distribution.
Therefore, it can be considered as the amount of error that naturally occurs in the estimates of the target variable. 

\textbf{F1 Score:}\label{F1-score}
%
By introducing a threshold value $b$, it is possible to gauge how many of the action predictions are considered correct and thereby obtain an evaluation of the binary classifier.
%
%
The threshold $b$ is selected using domain knowledge. 
With the knowledge where the expected value for either the nominal or pattern error behaviour is, we can compare our predictions with the ground truth.
%
Our reason for including the F1 score in our composition metric is that it will be used to evaluate models within a real-world production environment. 
Within our target domain, a low false positive warning rate is required, as otherwise 
workers will consider warnings as unreliable and not trustworthy.
Given that alerts require investigation, false positives will result in a superfluous waste of time.

\section*{Empirical Evaluation} \label{sec:empirical_eval}

\subsection*{Experiment Setup}

For the empirical evaluation we 
first discuss the result of applying our 
error classification module to the dataset provided
by \replaced[id=JR]{Volkswagen Commercial Vehicles}{our  industry partner}.
This dataset contains hierarchical actions. 
However, to enhance our sequence-to-sequence model training we remove the hierarchy of actions to lessen the noise in the data.
Therefore, in the last data preprocessing step  we remove the hierarchy of actions, as superordinate actions document the total times of subactions.
%
We focus on a single station to test the hypothesis that pattern errors can be learnt from the completion time of actions within an action sequence. 
We consider an exemplary station that has 22 actions.
This workstation is of particular interest for \replaced[id=JR]{Volkswagen Commercial Vehicles}{our
industry partner}, as delays are frequently observed. 
%
For our error classification module we set the global threshold as ten times $d^a_{globalmax} = 10 \times d^a_{max}$.
For the scalar for obtaining $d^a_{globalmax}$ we ran preliminary trials with 3, 5, 10, but found the 
former two removed a large proportion of data
points, impacting the accuracy of the predictions of the seq2seq models.
We therefore chose a scalar of 10, allowing us to retain \replaced[id=JR]{94.8\%}{97\%} of the data points. 
The parameters chosen for our seq2seq models can be found in Table~\ref{tab:hyperparams}.
%
%
Four different seq2seq architectures $n$-$m$ are compared with respect to length of the input sequence $n$ and the number of outputs $m$: 5--2, 5--5, 5--7, 7--7.
We conducted 10 training runs per model architecture, and the results in Table~\ref{tab:Model_results} are the averages from applying the models to our test data, using a 80\% training, 20\% test split.
For the evaluation we set the F1 threshold $b = 10\%$. 
%
\added[id=JR]{Our preliminary tests were also conducted with a $5\%$ and $20\%$ threshold. The evaluation with $20\%$ of different models is challenging due to too much separation uncertainty.
A threshold of $5\%$ becomes problematic for actions that last only a comparatively short time, since the noise in the time detection is larger than the targeted prediction quality.}
After considering only actions below $d^a_{max}$ and then calculate the RMSE from all of them we get $k = 5.14$.
%
%
\begin{table}
\begin{center}
\begin{tabular}{| c | c|} 
\hline
\multicolumn{2}{|c|}{\textbf{LSTM or GRU Model configurations}} \\
\hline
Nodes per layer & 100 \\ 
\hline
Layers & 4 \\ 
\hline
Dropout & 0.2 \\ 
\hline
\hline
\multicolumn{2}{|c|}{\textbf{Transformer Model Parameters}} \\
\hline
Number of heads & 2 \\
\hline
Head Size & 256 \\
\hline
Feed Forward Dimension & 1024 \\
\hline
Number of Transformer Blocks & 4 \\
\hline
MLP Units & 1024 \\
\hline
Dropout & 0.1 \\
\hline
\hline
\multicolumn{2}{|c|}{\textbf{General}} \\
\hline
Epochs & 50 \\
\hline
Batch Size & 128 \\
\hline
Optimizer & Adam \\
\hline
Learning rate & 0.001 \\
\hline
\end{tabular}
\caption{Hyperparameters}
\label{tab:hyperparams}
\end{center}
\end{table}

\subsection*{Error Classification Results} 









\replaced[id=GP]{Upon applying the error classification module to 
our dataset we first remove 2106 out of 40536 sequences that contain outliers~(5.2\%). 
Next, we apply our MLE based approach, finding that 3.94\% of samples sequences containing at least one source error (without \knockonerrors), 61.20\% containing \knockonerrors and 6.54\% containing both. With respect to normal and misc samples, we have 0.068\% only normal, 0.0902\% only misc, and 18.62\% \emph{only} misc and normal.}{Upon applying the error classification module to 
our dataset we obtain 9.48\% sequences containing source errors,
70.34\% \knockonerrors, 95.13\% misc sequences and 85.65\% normal sequences.}
%
\added[id=JR]{An analysis of the dataset following preprocessing reveals that \replaced[id=GP]{71.68}{75}\% of sequences contain at least one error.}
Therefore, surprisingly the majority of the sequences
contain either a source or \knockonerrors. 
As mentioned, during preliminary trials we also find that the 
small percentage of misc sequences can negatively impact the 
performance of the seq2seq models. 
We discuss this in more detail in the evaluation of our
seq2seq model results below. 



\subsection*{Sequence-to-Sequence Model Results}

In this section we shall first compare the results for the four different seq2seq architecture types based on length of the input sequences and predictions. 
Then we shall take a closer look at the impact of the choice for the TARMSE weighting factor $\tau$ for evaluating our models. 
An overview of the results obtained for each setting is
provided in Table~\ref{tab:Model_results}, 
where the balance between $TARMSE$ and $F1$ is $\tau = 0.5$.
Finally, we conduct an ablation study, showing the extent to which including misc samples impacts the performance of our seq2seq models. 

\textbf{Setup 5-2:} We first consider the results 
for training a seq2seq model to predict two future action durations
based on five historic actions (setup 5-2).
The TARMSE of the GRU and LSTM models is at $0.2 \pm 0.05$ and $0.22 \pm 0.08$ while the Transformer performs best with $0.41 \pm 0.01$.
Yet the summarized F1 score is lower at $0.8 \pm 0.01$ while the GRU and LSTM are better with $0.94 \pm 0.01$ or $0.95 \pm 0.01$.
Combined the CTA shows us that the GRU at $59.89 \pm 3.02$ and LSTM at $58.49 \pm 4.08$ are minimal worse \wrt mean than the Transformer at $60.55 \pm 0.76$. 
However, the standard deviation shows us that the Transformer is more consistent.

\textbf{Setup 5-5:} In the next setup 5-5 we see a similar behavior to the 5-2 setup.
The TARMSE is for the GRU and LSTM at $0.24 \pm 0.02$ and $0.23 \pm 0.01$ respectively, and for the Transformer it is $0.44 \pm 0.00$. The F1 is $0.94 \pm 0.02$ for the GRU, $0.93 \pm 0.01$ for the LSTM and $0.80 \pm 0.02$ for the Transformer.
The CTA shows that the Transformer is better with $61.69 \pm 0.85$ than GRU's $58.67 \pm 1.33$ and LSTM's $58.11 \pm 1.53$.

\textbf{Setup 7-5:} Next we keep the number of future predictions the same but consider a history of seven actions.
The TARMSE for GRU is $0.22 \pm 0.03$, LSTM is $0.20 \pm 0.04$ and Transformer slightly increasing than the previous 5-5 setup to now $0.49 \pm 0.01$.
The F1 score slightly decrease to $0.89 \pm 0.02$ for the GRU, $0.91 \pm 0.03$ for the LSTM and $0.81 \pm 0.01$ for the Transformer.
We notice a slight improvement in the CTA for the Transformer at $64.75 \pm 0.59$ while the GRU at $55.83 \pm 2.12$ and LSTM at $55.80 \pm 2.19$ decrease and notably the standard deviation is significantly higher now compared to the 5-5 setup.

\textbf{Setup 7-7:} Lastly we consider seven previous actions in a sequence and let the models predict seven actions into the future.
The TARMSE of the GRU and LSTM are both at $0.21 \pm 0.05$ and the for Transformer at $0.48 \pm 0.00$.
It should be noted that the standard deviation for the Transformer is considered that low that the rounding shows zero here.
The F1 is for the GRU at $0.88 \pm 0.02$, for the LSTM at $0.92 \pm 0.01$ and for the Transformer similar to before $0.80 \pm 0.01$.
For the GRU and LSTM the CTA are at $54.24 \pm 3.58$ and $56.22 \pm 2.55$ while the Transformer is at $63.88 \pm 0.70$.
Across all setups we can observe that the the Transformer shows better performance when predicting future actions by considering the TARMSE.
We see an improving trend in the TARMSE for the Transformer the more input actions are considered and prediction range increased.
However the F1 score is higher for the GRU and LSTM models. 

\begin{center}
\begin{table*}
\resizebox{\textwidth}{!}{
\begin{tabular}{|l|l|l|l|l|l|l|l|l|l|l|l|l|}
\hline
\textbf{Metric} & 
\textbf{GRU 5-2} & 
\textbf{LSTM 5-2} & 
\textbf{TF 5-2} &
\textbf{GRU 5-5} & 
\textbf{LSTM 5-5} & 
\textbf{TF 5-5} &
\textbf{GRU 7-5} & 
\textbf{LSTM 7-5} & 
\textbf{TF 7-5} &
\textbf{GRU 7-7} & 
\textbf{LSTM 7-7} & 
\textbf{TF 7-7} \\
\hline
$\mathbf{RMSE_1}$ & $4.14 \pm 0.24$ & $4.01 \pm 0.39$ & $2.95 \pm 0.02$ & $4.07 \pm 0.18$ & $4.03 \pm 0.21$ & $3.04 \pm 0.02$ & $3.99 \pm 0.22$ & $4.08 \pm 0.32$ & $2.68 \pm 0.04$ & $4.10 \pm 0.32$ & $4.05 \pm 0.40$ & $2.72 \pm 0.03$ \\
\hline
$\mathbf{RMSE_2}$ & $4.12 \pm 0.63$ & $4.06 \pm 0.49$ & $3.29 \pm 0.05$ &$3.72 \pm 0.29$ & $3.88 \pm 0.21$ & $2.70 \pm 0.02$ & $3.97 \pm 0.19$ & $4.15 \pm 0.35$ & $2.55 \pm 0.02$ & $4.00 \pm 0.29$ & $4.07 \pm 0.26$ & $2.56 \pm 0.02$ \\
\hline
$\mathbf{RMSE_3}$ & NA & NA & NA & $3.50 \pm 0.18$ & $3.84 \pm 0.30$ & $2.43 \pm 0.02$ & $4.09 \pm 0.41$ & $4.21 \pm 0.24$ & $2.59 \pm 0.02$ & $4.24 \pm 0.41$ & $4.30 \pm 0.16$ & $2.57 \pm 0.03$ \\
\hline
$\mathbf{RMSE_4}$ & NA & NA & NA & $3.51 \pm 0.23$ & $3.68 \pm 0.34$ & $2.35 \pm 0.02$ & $3.77 \pm 0.35$ & $3.92 \pm 0.34$ & $2.57 \pm 0.04$ & $4.06 \pm 0.28$ & $3.82 \pm 0.32$ & $2.57 \pm 0.02$ \\
\hline
$\mathbf{RMSE_5}$ & NA & NA & NA & $4.29 \pm 0.41$ & $4.09 \pm 0.53$ & $2.77 \pm 0.06$ & $4.23 \pm 0.52$ & $4.29 \pm 0.84$ & $2.78 \pm 0.10$ & $4.15 \pm 0.27$ & $4.15 \pm 0.29$ & $2.66 \pm 0.05$ \\
\hline
$\mathbf{RMSE_6}$ & NA & NA & NA & NA & NA & NA & NA & NA & NA & $3.49 \pm 0.31$ & $3.51 \pm 0.98$ & $2.86 \pm 0.06$ \\
\hline
$\mathbf{RMSE_7}$ & NA & NA & NA & NA & NA & NA & NA & NA & NA & $3.28 \pm 0.24$ & $3.68 \pm 0.56$ & $2.70 \pm 0.07$ \\
\hline
$\mathbf{F1_1}$ & $0.94 \pm 0.02$ & $0.95 \pm 0.01$ & $0.80 \pm 0.02$ & $0.94 \pm 0.02$ & $0.94 \pm 0.02$ & $0.79 \pm 0.03$ & $0.89 \pm 0.03$ & $0.91 \pm 0.04$ & $0.81 \pm 0.01$ & $0.87 \pm 0.04$ & $0.92 \pm 0.02$ & $0.80 \pm 0.01$ \\
\hline
$\mathbf{F1_2}$ & $0.95 \pm 0.01$ & $0.95 \pm 0.01$ & $0.82 \pm 0.03$ & $0.95 \pm 0.02$ & $0.94 \pm 0.01$ & $0.81 \pm 0.02$ & $0.89 \pm 0.02$ & $0.90 \pm 0.02$ & $0.81 \pm 0.02$ & $0.89 \pm 0.03$ & $0.90 \pm 0.01$ & $0.80 \pm 0.02$ \\
\hline
$\mathbf{F1_3}$ & NA & NA & NA & $0.89 \pm 0.03$ & $0.89 \pm 0.02$ & $0.79 \pm 0.03$ & $0.94 \pm 0.01$ & $0.94 \pm 0.01$ & $0.81 \pm 0.01$ & $0.91 \pm 0.02$ & $0.93 \pm 0.01$ & $0.79 \pm 0.01$ \\
\hline
$\mathbf{F1_4}$ & NA & NA & NA & $0.89 \pm 0.01$ & $0.90 \pm 0.01$ & $0.80 \pm 0.01$ & $0.94 \pm 0.01$ & $0.94 \pm 0.01$ & $0.81 \pm 0.01$ & $0.93 \pm 0.02$ & $0.94 \pm 0.01$ & $0.79 \pm 0.01$ \\
\hline
$\mathbf{F1_5}$ & NA & NA & NA & $0.92 \pm 0.01$ & $0.93 \pm 0.01$ & $0.78 \pm 0.02$ & $0.92 \pm 0.01$ & $0.92 \pm 0.03$ & $0.75 \pm 0.02$ & $0.85 \pm 0.04$ & $0.89 \pm 0.02$ & $0.75 \pm 0.02$ \\
\hline
$\mathbf{F1_6}$ & NA & NA & NA & NA & NA & NA &  NA & NA & NA & $0.85 \pm 0.03$ & $0.88 \pm 0.04$ & $0.72 \pm 0.02$ \\
\hline
$\mathbf{F1_7}$ & NA & NA & NA & NA & NA &  NA & NA & NA & NA & $0.85 \pm 0.04$ & $0.88 \pm 0.03$ & $0.76 \pm 0.03$ \\
\hline
$\mathbf{TARMSE}$ & $0.20 \pm 0.05$ & $0.22 \pm 0.08$ & $0.41 \pm 0.01$ & $0.24 \pm 0.02$ & $0.23 \pm 0.03$ & $0.44 \pm 0.00$ & $0.22 \pm 0.03$ & $0.20 \pm 0.04$ & $0.49 \pm 0.01$ & $0.21 \pm 0.05$ & $0.21 \pm 0.05$ & $0.48 \pm 0.00$ \\
\hline
$\mathbf{F1}$ & $0.94 \pm 0.01$ & $0.95 \pm 0.01$ & $0.80 \pm 0.01$ & $0.94 \pm 0.02$ & $0.93 \pm 0.01$ & $0.80 \pm 0.02$ & $0.89 \pm 0.02$ & $0.91 \pm 0.03$ & $0.81 \pm 0.01$ & $0.88 \pm 0.02$ & $0.92 \pm 0.01$ & $0.80 \pm 0.01$ \\
\hline
$\mathbf{CTA}$ \textbf{(\%)} & $59.89 \pm 3.02$ & $58.49 \pm 4.08$ & $60.55 \pm 0.76$ & $58.67 \pm 1.33$ & $58.11 \pm 1.53$ & $61.69 \pm 0.85$ & $55.83 \pm 2.12$ & $55.80 \pm 2.19$ & $64.75 \pm 0.59$ & $54.24 \pm 3.58$ & $56.22 \pm 2.55$ & $63.88 \pm 0.70$ \\
\hline
\end{tabular}}
\caption{Results table comparing each of our LSTM, GRU and Transformer (TF) architectures $n$-$m$, where $n$ represents the number of inputs, and $m$ the number of model outputs. The table provides RMSE and F1 scores for each outputs, as well as TARMSE, averge F1 and composite time-weighted actions (CTA) scores (using $\tau = 0.5$ for the later).}
\label{tab:Model_results}
\end{table*}
\end{center}

\begin{figure}[h]
    \centering
    \includegraphics[width=0.8\textwidth]{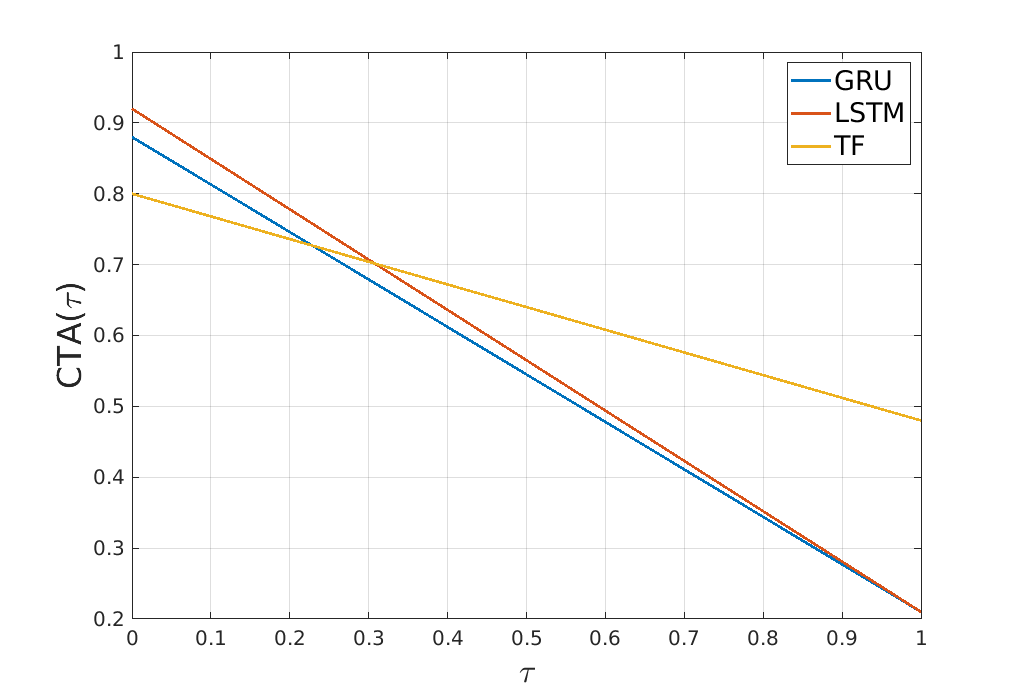}
    \caption{Impact of the weighting parameter $\tau$ for the composite time-weighted actions metric. Depicted are the results for GRU, LSTM and Transformer (TF) considering seven actions in the past and predicting seven future actions.}
    \label{CTA_weighting}
\end{figure}

\textbf{CTA Weighting Factor:} We note that the weighting factor $\tau$ influences our final result for the CTA.
In Figure \ref{CTA_weighting} we demonstrate the weighting factor between TARMSE and F1 for the chosen models in our setup with seven past actions to be considered and seven actions need to be predicted.
GRU and LSTM demonstrate here that due to their higher F1 score they initially start higher than the Transformer model.
With increasing $\tau$ the Transformer model surpasses the GRU model ($\tau = 0.229$) and LSTM model ($\tau = 0.308$) because of its better TARMSE.
%

\begin{figure}[h]
\centering
\includegraphics[width=0.8\textwidth]{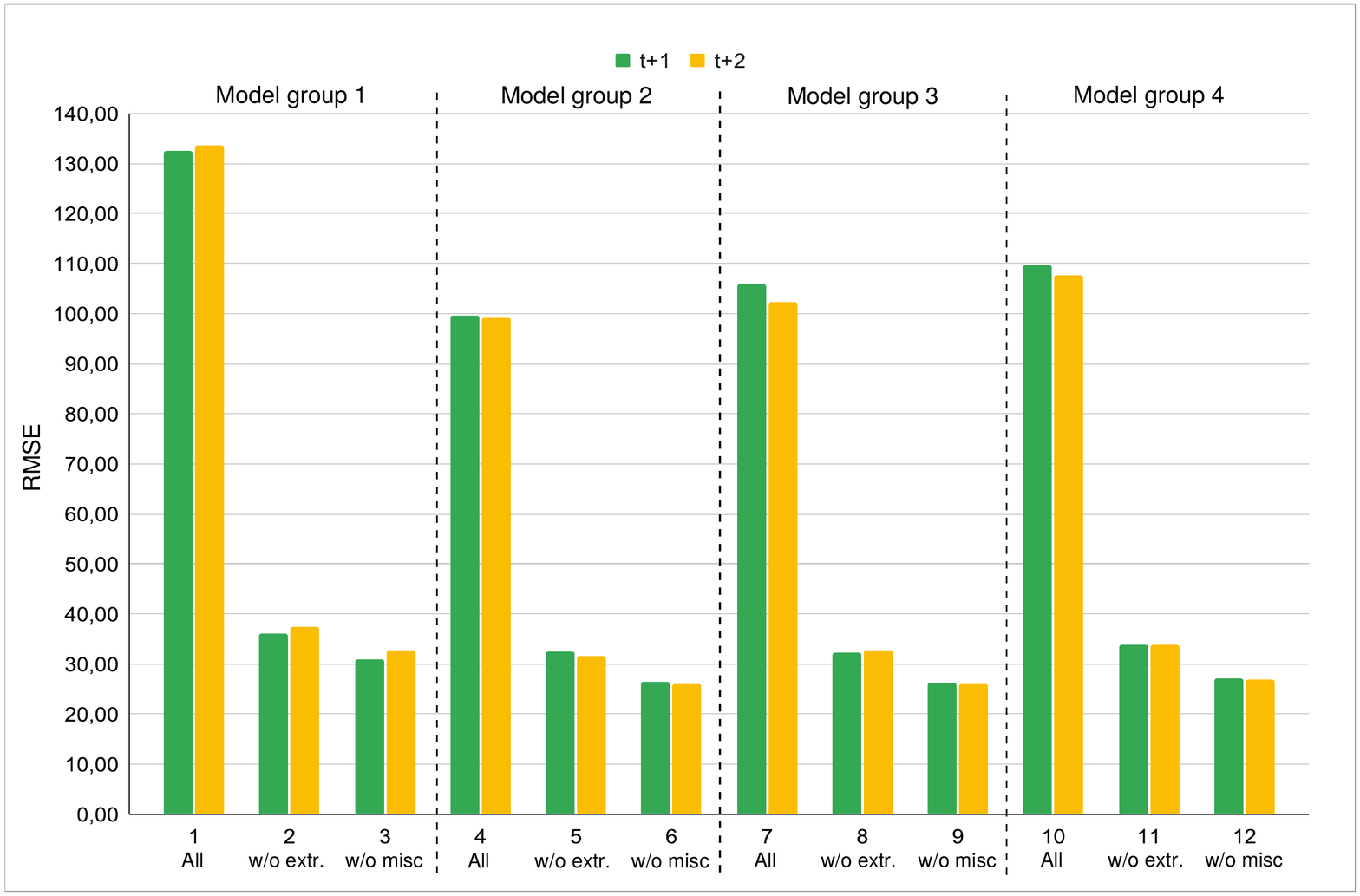}
\caption{RMSE effect of model performances including misc samples (1, 4, 7, 10), samples where the extreme outlier element were removed (2, 5, 8, 11) and misc samples completely removed (3, 6, 9, 12).}
\label{fig:ablation_study}
\end{figure}

\textbf{Ablation Study:}
As mentioned above, during preliminary trials we found that the inclusion of misc samples reduced the performance of the seq2seq models when used during training. 
We illustrate this in Figure~\ref{fig:ablation_study}, where we observe the RMSE of four model groups.
In each group the model is the same, but differ in the training and test set.
In each group the first experiment (1, 4, 7, 10) includes the misc samples.
The second of each group (2, 5, 8, 11) has \added[id=JR]{their} extreme element in the sequence removed, effectively skipping one process steps always.
The third of each group (3, 6, 9, 12) has the entire misc samples removed.
The exclusion of the extreme element in the misc samples improves the model performance by a factor three to four.
Since the removal of an extreme element in the misc samples \replaced[id=JR]{does not}{doesn't} mirror the real world application we opted to remove the entire sequence and achieve in average an additional $18\%$ model performance increase.

\section*{Future Work} \label{sec;manager_insights}

The results show that our VMAS can deliver interesting insights on real world data obtained from a PDA system for car manufacturing.
However, in order to measure the added value for stakeholders, the approach must be evaluated using key performance indicators.
Only in this way is it possible to derive optimization processes from the results in a targeted manner.
In practice, due to the extensive training time required for training seq2seq models, it makes sense 
to make use of components from our VMAS for a two-stage approach.

\noindent{\textbf{Stage 1:}} In a first integration stage, the results of the automatic peak detection and source error identification are used for the automatic identification of work steps which are particularly critical based on the frequency with which faults occur. 
Here, however, only a superficial analysis based the proportions of errors is possible. 
The deep-dive into the cause-effect relationships of the errors and, thus the identification of particularly critical faults, must still be done manually. 

\noindent{\textbf{Stage 2:}} Use a trained seq2seq model to automatically identify cause-effect relationships, and investigate which source faults actually result in the most disruption times and should therefore be eliminated first. 
Here, a measure such as the sum over all disturbance times would be required against which the each source error can be measured, to determine how critical it is. 
This would allow us to create a ranking, replacing the manual analysis from phase 1, after complete integration and successful training of the ML model.

\section*{Conclusion}


\added[id=GP]{In car body production, the car body is processed according to the order requirements at interlinked production stations.  
Frequently, faults are detected at stations, where the resulting disturbances not only affect the station itself, but also have a negative
impact on the downstream stations.
To address this problem we introduce a novel vehicle manufacturing analyses system that can identify the fault cause-effect relationships, and predict future delays. 
The evaluation of our framework on data from the car body production of \replaced[id=JR]{Volkswagen Commercial Vehicles}{a large German car manufacturer} shows that source and \knockonerrors are surprisingly prevalent, occurring in \replaced[id=GP]{71.68}{75}\% of action sequences.
Furthermore, we show that the prediction component of our model does 
well at predicting the durations of up to seven actions into the future, using state-of-the-art sequence-to-sequence models, including the Transformer. 
Therefore deployable framework can be used to efficiently process data for identifying source and \knockonerrors, as well as predicting future delays that can benefit from an early intervention.}

\bibliography{bibliography}



\section*{Acknowledgements}

The authors gratefully acknowledge, that the proposed research is a result of the
research project “IIP-Ecosphere”, granted by the German Federal Ministry for
Economics and Climate Action (BMWK) via funding code 01MK20006A. 

\section*{Author contributions statement}
All authors provided critical feedback and helped shape the research. 
Y.W. and J.R. contributed to the development and evaluation of the proposed methodology. 
Y.W. and S.L. contributed towards the development of the evaluation metrics. 
J.U. provided invaluable domain knowledge and helped formulate the research problem. 
J.R., S.W., and G.P. equally conceived the original idea and experimental design. 
S.W. and G.P. coordinated and supervised the project.
All authors reviewed the manuscript.
\end{document}